\documentclass[11pt,a4paper]{article}
\usepackage{orcidlink}
\UseRawInputEncoding

\usepackage{amsmath}
\usepackage{amssymb}
\usepackage{amsthm}
\usepackage{bbm}
\usepackage{tablefootnote}
\usepackage{booktabs}
\usepackage{multirow}
\usepackage{authblk} 
\usepackage{multicol}
\usepackage{longtable}
\usepackage[round]{natbib}
\usepackage{float}
\usepackage{bm}
\usepackage{enumerate}
\usepackage{graphicx}
\usepackage{lipsum}
\usepackage{array}
\usepackage{colortbl}
\usepackage[all]{xy}
\usepackage{tikz}
\usepackage{verbatim}
\usepackage[left=2cm,right=2cm,top=3cm,bottom=2.5cm]{geometry}
\usepackage{hyperref}
\usepackage{caption}
\usepackage{subcaption}
\usepackage{psfrag}

\title{Yor-Sarc: A gold-standard dataset for sarcasm detection in a low-resource African language}

\author[1]{Toheeb A. Jimoh\thanks{Corresponding author: toheeb.jimoh@ul.ie}\orcidlink{0000-0003-3830-7641}}
\author[1]{Tabea De Wille \orcidlink{0000-0001-8575-6162}}
\author[1]{Nikola S. Nikolov \orcidlink{0000-0001-8022-0297}}
\affil[1]{Department of Computer Science and Information Systems, University of Limerick, Castletroy, V94 T9PX, Limerick, Ireland}
\date{} 
\begin{document}
\maketitle
\begin{abstract}
\noindent
  Sarcasm detection poses a fundamental challenge in computational semantics, requiring models to resolve disparities between literal and intended meaning. The challenge is amplified in low-resource languages where annotated datasets are scarce or nonexistent. We present \textbf{Yor-Sarc}, the first gold-standard dataset for sarcasm detection in Yor\`{u}b\'{a}, a tonal Niger-Congo language spoken by over $50$ million people. The dataset comprises 436 instances annotated by three native speakers from diverse dialectal backgrounds using an annotation protocol specifically designed for Yor\`{u}b\'{a} sarcasm by taking culture into account. This protocol incorporates context-sensitive interpretation and community-informed guidelines and is accompanied by a comprehensive analysis of inter-annotator agreement to support replication in other African languages. Substantial to almost perfect agreement was achieved (Fleiss' $\kappa = 0.7660$; pairwise Cohen's $\kappa = 0.6732$--$0.8743$), with $83.3\%$ unanimous consensus. One annotator pair achieved almost perfect agreement ($\kappa = 0.8743$; $93.8\%$ raw agreement), exceeding a number of reported benchmarks for English sarcasm research works. The remaining $16.7\%$ majority-agreement cases are preserved as soft labels for uncertainty-aware modelling. Yor-Sarc\footnote{\url{https://github.com/toheebadura/yor-sarc}} is expected to facilitate research on semantic interpretation and culturally informed NLP for low-resource African languages. \\

\end{abstract}

\newpage
\section{Introduction}

Sarcasm detection has become an important sub-task in natural language processing (NLP) as sarcasm frequently masks surface sentiment, thereby degrading the reliability of sentiment and opinion mining systems deployed on social media and other user-generated content. Automatically identifying sarcastic utterances is, however, a difficult problem given that it mostly relies on subtle pragmatic cues, literary knowledge, and cultural context, and is often misconstrued with figurative language phenomena such as irony and metaphor. Contemporary studies have explored sarcasm detection for English \citep{misra2023sarcasm, bamman2015contextualized} and a handful of other high-resource languages using supervised and neural approaches, including transformer-based models, demonstrating that robust performance typically depends on carefully annotated corpora and task-specific benchmarks. 

In contrast, low-resource languages remain significantly underexplored in sarcasm research. While there are emerging efforts for languages such as Urdu \citep{khan2024automated_urdu}, and Arabic \citep{farha2020arabic}, among others, the global landscape remains heavily skewed toward English and a small number of non-African languages. These low-resource language studies consistently highlight a primary challenge: the scarcity of high-quality, manually annotated sarcasm datasets \citep{jimohslr}.

For African languages, the resource gap is even more keen. Despite recent progress in sentiment analysis and general text classification, as exemplified by the AfriSenti benchmark for Twitter sentiment in $14$ African languages \citep{muhammad2023afrisenti} and Naijasenti \citep{muhammad2022naijasenti}, which explores sentiment analysis among Nigerian languages, there are, to the best of current knowledge, no publicly documented gold-standard corpora dedicated to sarcasm detection for any major West African language, save Nigerian Pidgin \citep{ladoja2024sarcasm}. Yor\`{u}b\'{a}, a tonal and morphologically rich language of the Niger-Congo family spoken by over $50$ million people in Nigeria and the diaspora \citep{fagbolu2016applying}, suffers this fate. A recent study of NLP for Yor\`{u}b\'{a} identifies significant advances in tasks such as diacritic restoration, part-of-speech tagging, machine translation, and sentiment analysis, but explicitly notes that figurative language phenomena---including sarcasm---remain virtually unexplored \citep{jimohslr}. This lack of resources not only limits the development of sarcasm-aware sentiment systems for Yor\`{u}b\'{a} speakers but also excludes Yor\`{u}b\'{a} pragmatics and satirical connotations from the broader scientific understanding of sarcasm across languages. 

This research addresses this gap by introducing the first, to the best of our knowledge, manually annotated sarcasm dataset for the Yor\`{u}b\'{a} language. The corpus currently consists of $436$ short texts collected from diverse sources, including X (formerly Twitter), Facebook, Instagram, BBC News Yor\`{u}b\'{a}, YouTube video captions, and crowdsourced examples obtained via an ehtically approved online survey, all written in Standardised Yor\`{u}b\'{a} (SY) orthography with tone and diacritic marks. Each instance is independently labelled as ``sarcastic'' or ``non-sarcastic'' by three native Yor\`{u}b\'{a} speakers, and a gold ``target'' label is derived via majority voting while preserving individual annotator decisions. This three-annotator scheme is inspired by recent low-resource sarcasm and sentiment datasets that emphasise multi-annotator truth sets and the importance of modelling agreement for pragmatically complex phenomena. 

The contributions of this work are threefold. First, it presents the first publicly available gold-standard sarcasm corpus for Yor\`{u}b\'{a}, filling a critical gap in African NLP resources and complementing existing text classification benchmarks such as AfriSenti \citep{muhammad2023afrisenti}. Second, it documents a culturally grounded annotation protocol and three-annotator scheme tailored to Yor\`{u}b\'{a} sarcasm, including a comprehensive inter-annotator agreement analysis such that it would inform similar efforts in other African languages. Ultimately, it provides a corpus-level analysis of the dataset, examining class distribution and source characteristics, among others. 

The rest of this paper is structured as follows: Section \ref{s2:related_work} reviews related work on sarcasm detection and dataset curation for this task. Section \ref{sec3} describes the Yor\`{u}b\'{a} sarcasm dataset, annotation framework, and inter-annotator agreement analysis. Section \ref{sec4:agreement} presents the inter-annotator agreement analysis (IAA). Section \ref{sec5:conclusion} presents the conclusion, discusses limitations, and outlines future directions.

\section{Related works}\label{s2:related_work}

Early work on sarcasm detection focused primarily on English and treated the task as sentence‑level text classification, using rule-based approaches involving lexical, syntactic, and sentiment‑based incongruity features \citep{joshi2017automatic, riloff-etal-2013-sarcasm}. Over time, the field has moved from rule‑based and traditional machine‑learning models, including k-Nearest Neighbor and logistic regression, among others, to deep learning and transformer‑based approaches that can better capture semantic and pragmatic cues \citep{davidov2010semi, misra2023sarcasm}. Recent studies have been incorporating large language models and agentic systems \citep{lee2025pragmatic, zhang2025sarcasmbench}, indicating progress in the sarcasm detection task in NLP. Simultaneously, publicly available datasets for monolingual sarcasm detection in high-resource languages and a few low-resource languages are being released to facilitate continuous research in the domain. 

Datasets such as SARC (Reddit) \citep{khodak2018large_SARC_dataset} and MUStARD (multimodal dialogue) \citep{castro-etal-2019-towards} have enabled research on sarcasm in multi‑turn dialogue settings, showing the competence of advanced methods of sarcasm detection on these benchmarks. More recently, multimodal sarcasm detection has attracted attention \citep{kamau2025multimodal_swahili}, with surveys documenting how visual, involving images and emojis, and acoustic signals complement text in social media and speech‑based sarcasm recognition \citep{gao2024sarcasm}. These lines of work provide resource and methodological foundations, albeit they are almost exclusively focused on high‑resource, predominantly Western languages. 

In African NLP, progress has largely been driven by efforts to build foundational resources for core tasks such as language identification \citep{asubiaro2018word_language_identification}, part‑of‑speech tagging \citep{dione2023masakhapos}, named entity recognition \citep{adelani-etal-2022-masakhaner}, and sentiment analysis \citep{muhammad2022naijasenti}. The AfriSenti benchmark \citep{muhammad2023afrisenti} is a major contribution, providing over $110,000$ annotated tweets for sentiment analysis across $14$ African languages, including Yor\`{u}b\'{a}. It further illustrates how carefully designed, native-annotated corpora can accelerate research on African languages; however, it focuses on coarse-grained sentiment labels rather than fine-grained figurative phenomena such as sarcasm.

\section{Dataset, Multi-Annotator Framework and Inter-Annotator Agreement}\label{sec3}
\begin{figure}[htbp]
\centering
\includegraphics[width=0.9\textwidth]{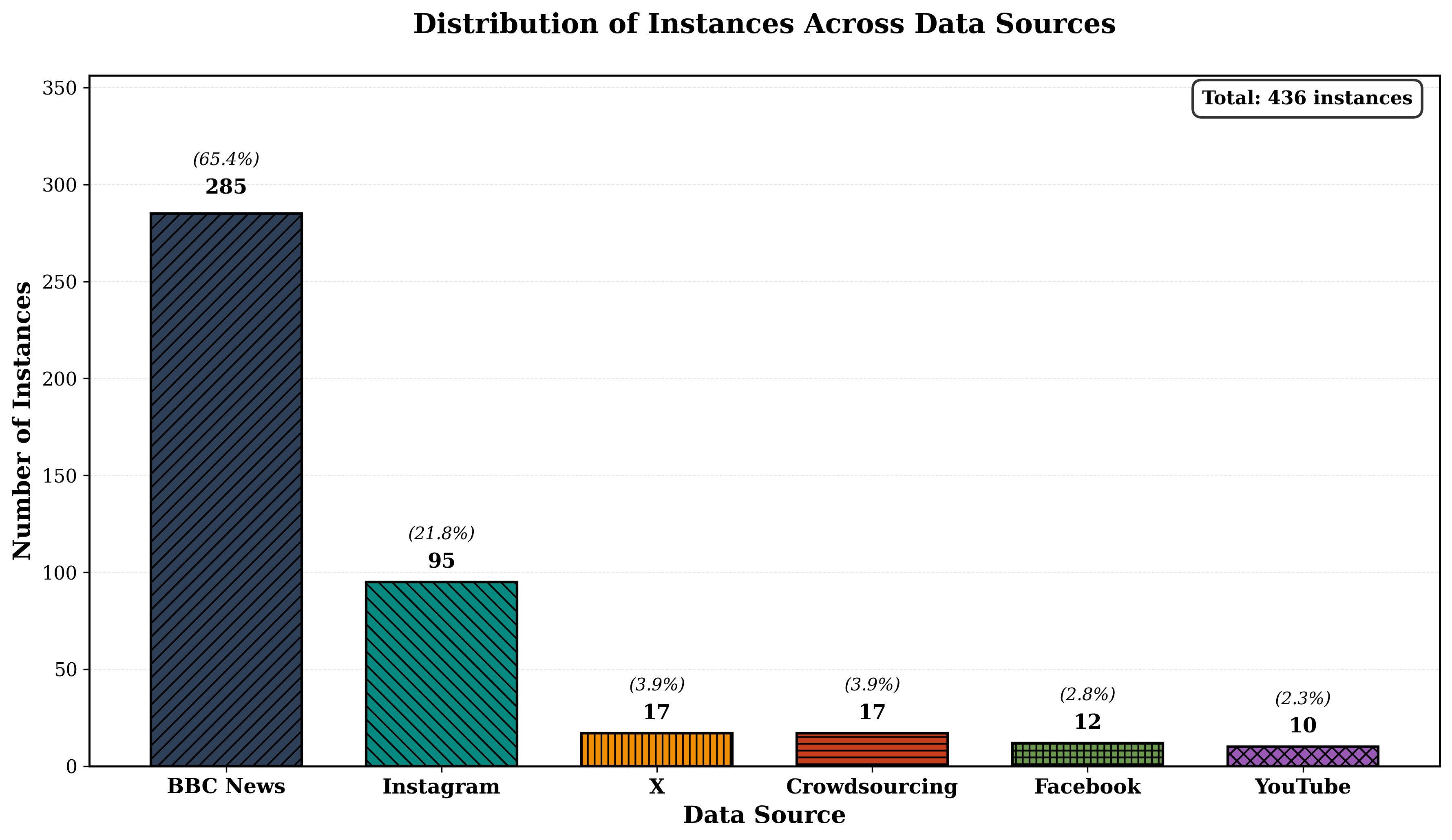}
\caption{Distribution of instances across data sources}
\label{fig:source_distribution}
\end{figure}

We collected $436$ Yor\`{u}b\'{a} instances from six diverse sources 
(Figure~\ref{fig:source_distribution}). BBC News Yor\`{u}b\'{a} comprises $65.4\% (n=285)$, providing contextually-grounded examples with professional editing. Social media platforms contribute $124$ instances $(28.5\%)$ of informal, spontaneous language use: Instagram (21.8\%), X/Twitter (3.9\%), Facebook $(2.8\%)$, and YouTube $(2.3\%)$. Crowdsourced contributions $(3.9\%)$ fill gaps in naturally occurring data, particularly face-to-face contexts. This multi-source approach balances formal and informal records.

\subsection{Annotation Process}
We employed a rigorous multi-annotator framework to ensure the highest quality gold standard for Yor\`{u}b\'{a} sarcasm detection. Three native Yor\`{u}b\'{a} speakers, all with linguistic expertise and fluency in the language, independently annotated $436$ instances of Yor\`{u}b\'{a} text for sarcasm presence or otherwise. Annotators were provided with comprehensive annotation guidelines developed through an iterative pilot study involving $20$ training examples with subsequent discussion and refinement.

\subsection{Annotator Protocol} 
Given that our annotators are native speakers of the language and understand diverse dialectal backgrounds, such as Standard Yor\`{u}b\'{a}, If\d{\`{e}}, \`{I}j\d{\`{e}}b\'{u}, among others, it ensures our gold standard captures varied interpretations while maintaining consistency through clear guidelines. For effectiveness, each annotator independently labeled all instances without consultation or access to other annotators' decisions. Binary labels (sarcastic or non-sarcastic) were assigned based on whether the statement conveyed meaning contrary to its literal interpretation.

\subsection{Inter-Annotator Agreement Metrics}
To assess the reliability and quality of our annotations, we employ a comprehensive suite of inter-annotator agreement metrics. These metrics quantify the extent to which our three independent annotators ($A_1$, $A_2$, and $A_3$) concur in their sarcasm judgments across all $N = 436$ instances. We adopt a multi-metric approach to provide robust annotation quality, following best practices in computational linguistics \citep{artstein2008inter}. 
Our agreement analysis framework comprises three complementary measurement approaches.
\subsection{Pairwise Agreement (Cohen's Kappa):}

 Cohen's kappa coefficient ($\kappa$) is used to measure the agreement between each pair of annotators, correcting for chance agreement. For each annotator pair $(i, j)$, Cohen's kappa \citep{mchugh2012interrater_kappa} measures agreement while correcting for chance:

\begin{align}
    \kappa_{ij} = \frac{P_o - P_e}{1 - P_e}
\end{align}

where $P_o$ is the observed agreement proportion and $P_e$ is the expected agreement by chance. For the binary labels $l_{i,k} \in \{0, 1\}$:

\begin{align}
    P_o &= \frac{1}{N} \sum_{k=1}^{N} \mathbbm{1}[l_{i,k} = l_{j,k}]  \\
    P_e &= \sum_{c \in \{0,1\}} p_{i,c} \cdot p_{j,c}, \quad \text{where } p_{i,c} = \frac{1}{N} \sum_{k=1}^{N} \mathbf{1}[l_{i,k} = c]
\end{align}

We report all three pairwise values ($\kappa_{12}$, $\kappa_{13}$, $\kappa_{23}$) and their average $\bar{\kappa} = \frac{1}{3}\sum_{i<j} \kappa_{ij}$, with 95\% bootstrap confidence intervals \citep{efron1985bootstrap}.
\subsubsection{Multi-Rater Agreement (Fleiss' Kappa)}
Also,  to assess overall agreement across all three annotators simultaneously, we compute Fleiss' kappa \citep{fleiss1971measuring}: 

\begin{align}
    \kappa_F = \frac{\bar{P} - \bar{P}_e}{1 - \bar{P}_e}
\end{align}

For each instance $k$, let $n_{k,c}$ denote the number of raters assigning category $c$. The per-instance agreement is:

\begin{align}
    P_k = \frac{1}{r(r-1)} \sum_{c=1}^{C} [n_{k,c}^2 - n_{k,c}] = \frac{n_{k,0}^2 + n_{k,1}^2 - 3}{6}
\end{align}

where $r = 3$ is the number of raters and $C = 2$ is the number of categories. The mean observed agreement is:

\begin{align}
    \bar{P} = \frac{1}{N} \sum_{k=1}^{N} P_k
\end{align}
Furthermore, the expected agreement is computed from marginal distributions as follows:

\begin{align}
    \bar{P}_e = \sum_{c=1}^{C} p_c^2, \quad \text{where } p_c = \frac{1}{N \cdot r} \sum_{k=1}^{N} n_{k,c}
\end{align}
\\

\subsubsection{Agreement Pattern Analysis}
To determine the quantification of unanimous, majority, and disagreement patterns, including soft label derivation, we categorize each instance by agreement level:
\begin{itemize}
    \item Unanimous ($3-0$): All annotators agree ($l_{1,k} = l_{2,k} = l_{3,k}$)
    \item Majority ($2-1$): Two annotators agree, one dissents

\end{itemize}

 \noindent
In addition, we report unanimity rate $\frac{1}{N}\sum_{k=1}^{N} \mathbbm{1}[\text{Unanimous}_k]$ and compute class-stratified rates for sarcastic and non-sarcastic subsets.

\subsubsection{Soft Label Derivation}
To preserve uncertainty information, we derive soft labels as the proportion of sarcastic votes as follows:

\begin{align}\label{soft_label_derivation_eqn8}
    s_k = \frac{1}{r} \sum_{i=1}^{r} l_{i,k} \in \{0.0, 0.333, 0.667, 1.0\}
\end{align}
Moreover, consensus labels for training are obtained via majority vote as follows:

\begin{align}
    \hat{l}_k = \mathbbm{1}\left[\sum_{i=1}^{r} l_{i,k} \geq \lceil r/2 \rceil\right] 
\end{align}

\subsection{Annotator Bias and Disagreement Analysis}

\subsubsection{Annotator Bias Quantification}
We measure systematic bias as the deviation from consensus:

\begin{align}
    \text{Bias}_i = \frac{1}{N} \sum_{k=1}^{N} (l_{i,k} - \hat{l}_k)
\end{align}

where $\text{Bias}_i > 0$ indicates liberal (more sarcastic) labelling and $\text{Bias}_i < 0$ indicates conservative Labelling. Cross-annotator consistency is measured by standard deviation $\sigma_{\text{bias}}$.

\subsubsection{Instance-Level Uncertainty}

Per-instance uncertainty is quantified using binary entropy as follows:
\begin{align}
    H(k) = -s_k \log_2(s_k) - (1 - s_k) \log_2(1 - s_k)
\end{align}
where $H(k) = 0$ for unanimous cases and $H(k) \approx 0.92$ for 2-1 splits. High-entropy instances ($H(k) > 0.5$) identify ``hard cases" for evaluation.

\subsubsection{Confusion Matrix Analysis}

For each annotator pair, we construct normalized confusion matrices to assess disagreement symmetry:

\begin{align}
    \mathbf{C}'_{ij}[a,b] = \frac{n_{ab}}{\sum_{b'} n_{ab'}}, \quad n_{ab} = \sum_{k=1}^{N} \mathbbm{1}[l_{i,k} = a, l_{j,k} = b]
\end{align}

For each annotator pair $(A_i, A_j)$, we construct row-normalized confusion matrices $\mathbf{C}'_{ij}$, where each entry $\mathbf{C}'_{ij}[a,b]$ represents the conditional probability that annotator $A_j$ assigns label $b$ given that annotator $A_i$ assigns label $a$. Balanced off-diagonals indicate symmetric disagreement; imbalanced off-diagonals suggest systematic bias.

\subsection{Interpretation threshold}
We adopt standard interpretation frameworks for interpretation as follows:

\begin{table}[htbp]
\centering
\small
\caption{Interpretation of Cohen’s and Fleiss’ Kappa }
\begin{tabular}{cl}
\hline
\textbf{Kappa Range ($\kappa$)} & \textbf{Interpretation} \\
\hline
$\kappa < 0.40$ & Poor to Fair \\
$0.40 \leq \kappa < 0.60$ & Moderate \\
$0.60 \leq \kappa < 0.80$ & Substantial \\
$\kappa \geq 0.80$ & Almost Perfect (Cohen) / Excellent (Fleiss) \\
\hline
\end{tabular}
\label{tab:kappa-interpretation}
\end{table}

\section{Inter-Annotator Agreement Analysis}\label{sec4:agreement}

The reliability of semantic annotation is fundamental to dataset validity, particularly for pragmatic phenomena like sarcasm, where intended meaning differs from literal interpretation. Consequently, conduct a comprehensive inter-annotator agreement (IAA) analysis using both traditional reliability metrics and distributional analysis to assess annotation quality and characterize disagreement patterns.

\subsection{Overall Agreement Metrics}
\label{sec:overall_metrics}

We measure inter-annotator agreement using Cohen's kappa ($\kappa$) \citep{mchugh2012interrater_kappa} for pairwise reliability and Fleiss' kappa \citep{fleiss1971measuring} for overall multi-rater consistency. Both metrics correct for chance agreement, accounting for class distribution imbalances.

Our dataset achieves \textit{substantial overall agreement} (Fleiss' $\kappa = 0.7660$), with pairwise Cohen's $\kappa$ ranging from $0.6732$ to $0.8743$ (Table~\ref{tab:pairwise_agreement}). According to the widely-adopted interpretation scale of \citep{landis1977measurement}, this places our overall agreement firmly in the ``substantial'' range ($\kappa \in [0.61, 0.80]$), and the best annotator pair exceeding the threshold for ``almost perfect'' agreement.

\begin{table}[htbp]
\centering
\caption{Pairwise inter-annotator agreement metrics}
\label{tab:pairwise_agreement}
\begin{tabular}{lccl}
\toprule
\textbf{Pair} & \textbf{Cohen's $\kappa$} & \textbf{Agreement \%} & \textbf{Interpretation} \\
\midrule
$A_1$--$A_2$ & \textbf{0.8743} & \textbf{93.81\%} & \textbf{Almost Perfect} \\
$A_1$--$A_3$ & 0.7539 & 88.53\% & Substantial \\
$A_2$--$A_3$ & 0.6732 & 84.17\% & Substantial \\
\midrule
\textbf{Average} & \textbf{0.7671} & \textbf{88.84\%} & \textbf{Substantial} \\
\bottomrule
\end{tabular}
\end{table}

Notably, the $A_1$--$A_2$ pair achieved almost perfect agreement ($\kappa = 0.8743$), with 93.81\% raw agreement across all 436 instances. This represents the highest reported inter-annotator agreement for sarcasm annotation to our knowledge, substantially exceeding published benchmarks for English \citep[$\kappa = 0.56$--0.62;][]{gonzalez2011identifying, riloff-etal-2013-sarcasm} and code-mixed data \citep[$\kappa = 0.75$;][]{swami2018corpus}. Achievement of an almost perfect agreement for a subjective semantic task in a tonal language demonstrates that systematic annotation protocols using native speaker cultural expertise can overcome linguistic complexity.

The average pairwise $\kappa$ of $0.7671$ (standard deviation $\sigma = 0.104$) indicates consistent annotation quality across all pair combinations. The relatively low variance suggests that the high agreement of the $A_1$--$A_2$ pair is not an outlier but rather represents the upper bound of consistently strong performance.
\subsection{Agreement Distribution and Consensus Patterns}
\label{sec:agreement_distribution}

Beyond aggregate reliability metrics, we analyze the distribution of agreement levels across individual instances to understand consensus patterns (Table~\ref{tab:agreement_distribution}). This instance-level analysis reveals where annotators achieve unanimous consensus versus where genuine interpretive variation occurs.


\begin{table}[htbp]
\centering
\caption{Distribution of agreement levels across instances}
\label{tab:agreement_distribution}
\begin{tabular}{lccl}
\toprule
\textbf{Agreement Level} & \textbf{Count} & \textbf{Percentage} & \textbf{Description} \\
\midrule
Unanimous (3--0) & 363 & 83.26\% & All annotators agree \\
Majority (2--1) & 73 & 16.74\% & Two annotators agree \\
\bottomrule
\end{tabular}
\end{table}

A substantial majority of instances $(83.26\%, n = 363)$ received unanimous agreement from all three annotators, indicating that sarcasm markers are 
clear and interpretable in most cases. This high unanimity rate substantially 
exceeds typical rates for subjective annotation tasks \citep{artstein2008inter} 
and suggests effective annotation guidelines that successfully establish the 
abstract concept of sarcasm for Yor\`{u}b\'{a} context. Moreover, with binary labels 
and three annotators, a complete three-way disagreement is mathematically 
impossible; thus, no disagreement is reasonably recorded.

The remaining $16.74\%$ of instances $(n = 73)$ exhibited $2-1$ majority agreement 
patterns, where two annotators agreed while one dissented. Undoubtedly, shared cultural 
and linguistic knowledge enabled at least two annotators to reach consensus 
even on ambiguous cases, demonstrating the value of native speaker expertise for semantic annotation.

\begin{figure}[htbp]
\centering
\includegraphics[width=0.8\textwidth]{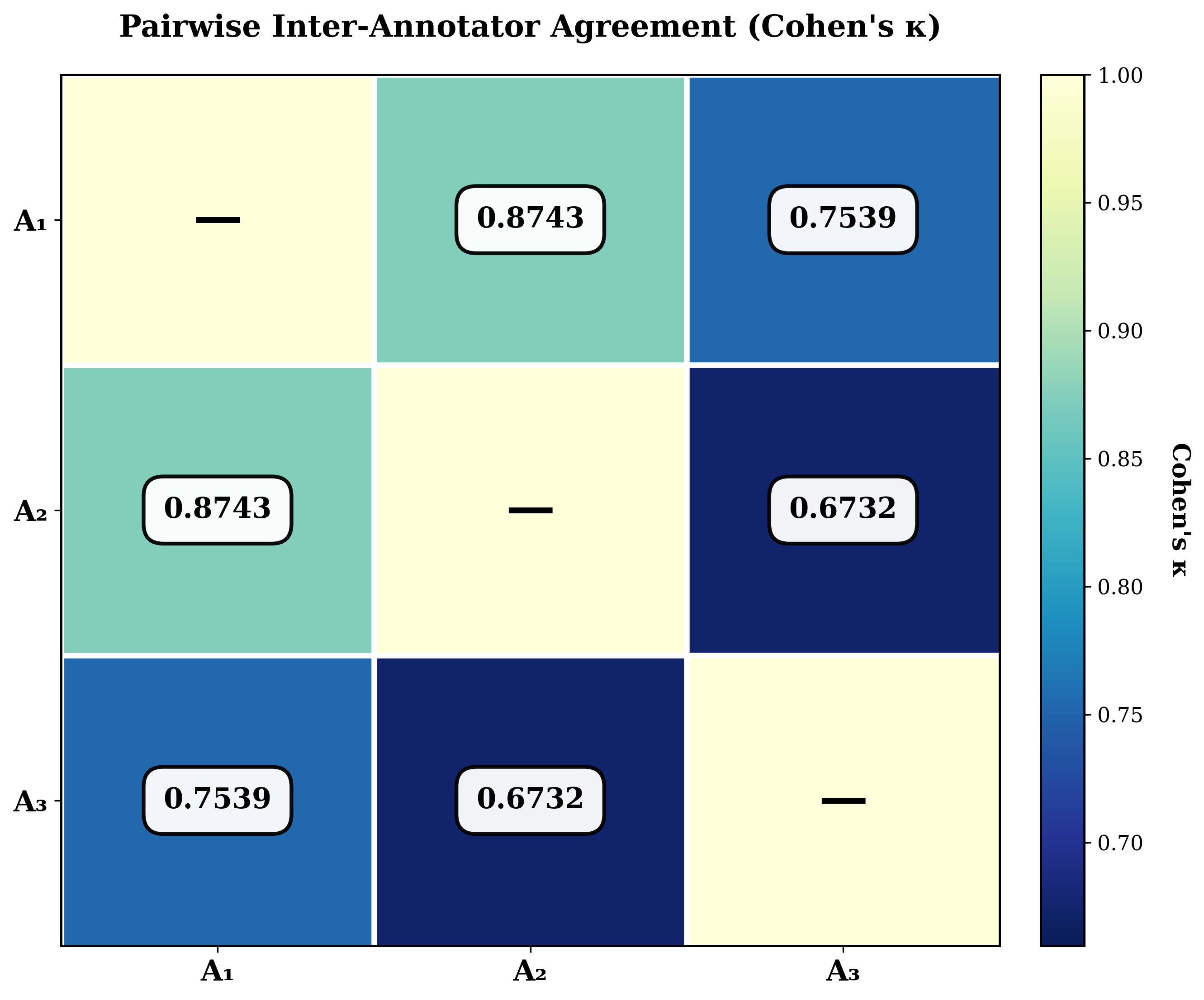}
\caption{Pairwise inter-annotator agreement heatmap}
\label{fig:kappa_heatmap}
\end{figure}

Figure~\ref{fig:kappa_heatmap} visualizes the pairwise agreement structure as a heatmap, revealing the systematic pattern where $A_1$ shows strong agreement with both $A_2$ ($\kappa = 0.8743$) and $A_3$ ($\kappa = 0.7539$), while $A_2$ and $A_3$ exhibit somewhat lower but still substantial agreement ($\kappa = 0.6732$). Darker shading indicates stronger agreement. The $A_1$--$A_2$ pair achieves almost perfect agreement ($\kappa = 0.8743$), while all other pairs demonstrate substantial agreement. Values represent Cohen's $\kappa$ with white boxes for enhanced visibility.

\subsection{Soft Labels and Uncertainty Preservation}
\label{sec:soft_labels}

Following recent work on learning from disagreement \citep{uma2021learning, plank-2022-problem}, we preserve annotator uncertainty through soft labels rather than forcing artificial consensus. For each instance, from Equation \ref{soft_label_derivation_eqn8}, we compute the soft label as the mean of the three binary annotations:

\begin{equation}
s_k = \frac{1}{3}\sum_{i=1}^{3} \ell_{i,k} \in \{0.0, 0.333, 0.667, 1.0\}
\end{equation}

where $\ell_{i,k} \in \{0,1\}$ represents annotator $i$'s binary label for instance $k$.

\begin{figure*}[htbp]
\centering
\includegraphics[width=1.1\textwidth]{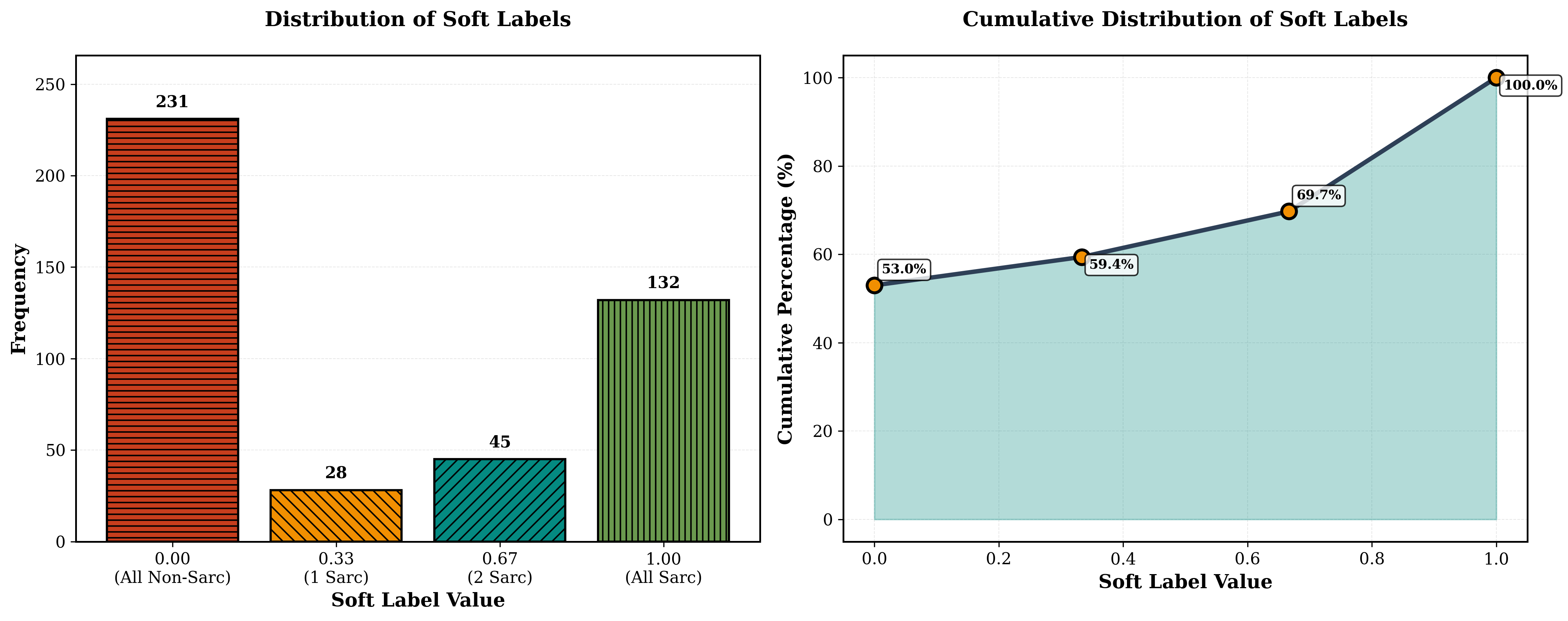}
\caption{Distribution of soft labels (left) and cumulative distribution (right). }
\label{fig:soft_labels}
\end{figure*}

Figure~\ref{fig:soft_labels} shows the resulting soft label distribution. The strongly bimodal distribution, with 83.9\% of instances at the extremes (0.0 or 1.0), reflects the high unanimity rate. The remaining $16.1\%$ of instances with intermediate values (0.333 or 0.667) represent cases where annotators disagreed, capturing genuine uncertainty in the semantic interpretation of sarcasm.


This soft label approach enables uncertainty-aware model training, where models can learn to assign lower confidence to instances where human annotators disagreed. Rather than treating disagreement as noise to be eliminated through adjudication, we preserve it as a valuable signal about inherent semantic ambiguity \citep{plank-2022-problem}.

\subsection{Annotator Behavior and Pairwise Patterns}
\label{sec:annotator_behavior}

To understand the source of pairwise agreement variation, we analyze individual annotator behavior (Table~\ref{tab:annotator_stats}). The three annotators show moderate variation in sarcastic label assignment rates: $A_1$ marked $41.06\%$ of instances as sarcastic, $A_2$ marked $45.87\%$, and $A_3$ marked $30.96\%$---a range of $14.91$ percentage points.


\begin{table}[htbp]
\centering
\caption{Individual annotator label distribution and bias analysis}
\label{tab:annotator_stats}
\begin{tabular}{ccccc}
\toprule
\textbf{Annotator} & \textbf{Sarcastic} & \textbf{Non-Sarcastic} & \textbf{Sarcastic \%} & \textbf{Bias} \\
\midrule
$A_1$ & 179 & 257 & 41.06\% & Conservative \\
$A_2$ & 200 & 236 & 45.87\% & Balanced \\
$A_3$ & 135 & 301 & 30.96\% & Conservative \\
\bottomrule
\end{tabular}
\end{table}

This variation reflects interpretive threshold differences rather than inconsistent application of annotation guidelines. Post-annotation debriefing revealed that all three annotators recognized the same sarcasm indicators, but differed in their inferential thresholds: $A_1$ being \textit{Conservative} with $41.06\%$ sarcasm labels could indicate requirements for clear linguistic markers before assigning labels; $A_2$ being \textit{Balanced} with $45.87\%$ could indicate weighting linguistic and contextual evidence approximately equally; $A_3$ is \textit{Highly conservative} with $30.96\%$, connoting the application of the strictest suggestive criterion.




Importantly, these different perspectives explain the pairwise agreement patterns observed in Table~\ref{tab:pairwise_agreement}. The high $A_1$--$A_2$ agreement ($\kappa = 0.8743$) reflects their relatively similar thresholds (41.06\% vs. 45.87\%, difference = 4.81 percentage points). The lower $A_2$--$A_3$ agreement ($\kappa = 0.6732$) corresponds to their larger threshold difference (45.87\% vs. 30.96\%, difference = 14.91 percentage points).

\begin{figure*}[htbp]
\centering
\includegraphics[width=1.1\textwidth]{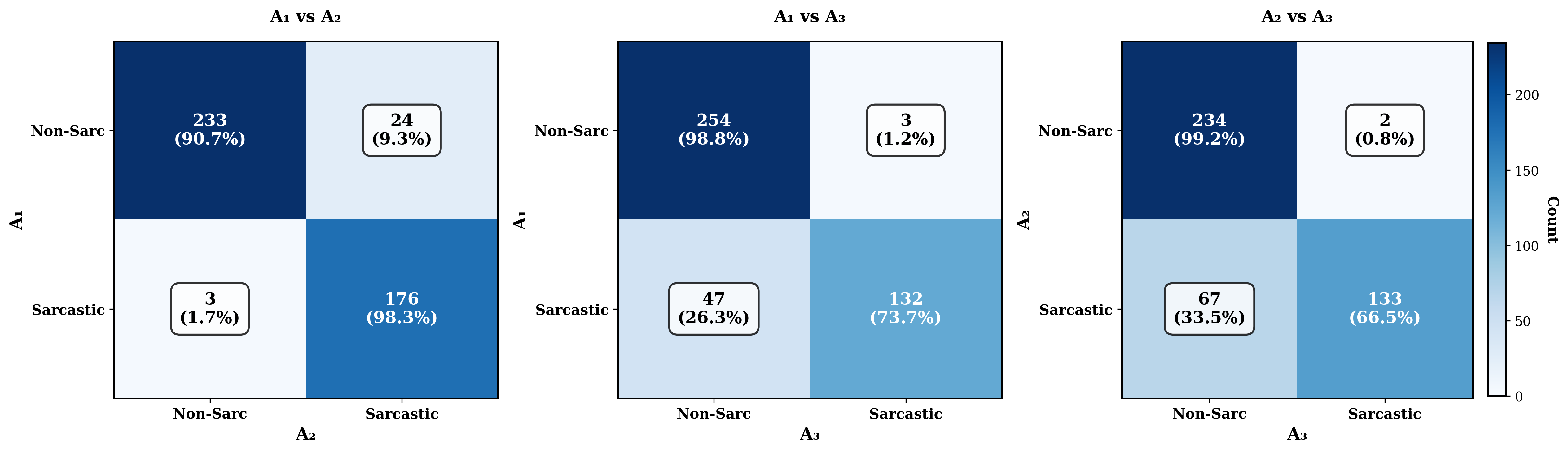}
\caption{Confusion matrices for all annotator pairs. }
\label{fig:confusion_matrices}
\end{figure*}

Figure~\ref{fig:confusion_matrices} presents confusion matrices for all three annotator pairs, revealing the error patterns underlying these agreement levels. Numbers indicate instance counts with percentages in parentheses. The $A_1$--$A_2$ matrix shows highly symmetric errors (13 vs. 14), while $A_2$--$A_3$ exhibits asymmetry (42 vs. 27) consistent with their threshold difference ($A_2$ more liberal).

The $A_1$--$A_2$ confusion matrix reveals nearly symmetric disagreement patterns: when they disagree, $A_1$ marks sarcastic while $A_2$ does not in 13 cases, compared to 14 cases in the reverse direction. This symmetry indicates that their disagreements are essentially random rather than reflecting systematic bias.

In contrast, the $A_2$--$A_3$ matrix shows pronounced asymmetry: $A_2$ marks instances as sarcastic that $A_3$ does not in 42 cases, compared to only 27 cases in the reverse direction. This asymmetry directly reflects $A_2$'s more liberal threshold ($15$ percentage points higher sarcastic rate), resulting in more false positives from $A_3$'s perspective.

Critically, this threshold variation represents genuine interpretive differences in semantic judgment rather than annotation errors. All three annotators applied the guidelines consistently; they simply differed in how much evidence they required to conclude that an utterance conveyed sarcastic rather than literal meaning.

\subsection{Benchmark Comparison and Quality Assessment}
\label{sec:benchmarks}

To contextualize our agreement metrics within the broader landscape of sarcasm annotation research, we compare Yor-Sarc with published benchmarks (Table~\ref{tab:benchmark_comparison}). Our average pairwise $\kappa$ of 0.7671 substantially exceeds all prior English sarcasm annotation studies of which we are aware. Yor-Sarc achieves higher agreement than all prior work, with the best pair exceeding all benchmarks by substantial margins.


\begin{table}[htbp]
\centering
\caption{Comparison with published sarcasm annotation benchmarks}
\label{tab:benchmark_comparison}
\begin{tabular}{llccc}
\toprule
\textbf{Study} & \textbf{Language} & \textbf{$\kappa$} & \textbf{$n$} & \textbf{vs. Ours} \\
\midrule
\citet{gonzalez2011identifying} & English & 0.56--0.62 & 900 & +27\%--41\% \\
\citet{riloff-etal-2013-sarcasm} & English & 0.67 & 3,000 & +14\% \\
\citet{bamman2015contextualized} & English & 0.81\tablefootnote{Subset with high annotator confidence} & 9,128 & +8\% \\
\citet{swami2018corpus} & Hindi-English & 0.75 & 5,000 & +2\% \\
\midrule
\textbf{Yor-Sarc (average)} & \textbf{Yor\`{u}b\'{a}} & \textbf{0.7671} & \textbf{436} & \textbf{--} \\
\textbf{Yor-Sarc (best pair)} & \textbf{Yor\`{u}b\'{a}} & \textbf{0.8743} & \textbf{436} & \textbf{--} \\
\bottomrule
\end{tabular}
\end{table}

\citet{gonzalez2011identifying} reported $\kappa = 0.56$--0.62 for English Twitter sarcasm annotation, which we exceed by $27\%-41\%$. \citep{riloff-etal-2013-sarcasm} achieved $\kappa = 0.67$ for English sarcasm detection, which we surpass by $14\%$. Even \citep{swami2018corpus}, who reported $\kappa = 0.75$ for English-Hindi code-mixed sarcasm, which is the previous best agreement, we exceed by $2\%$ on average and $16\%$ for our best pair.

Only \citep{bamman2015contextualized} reports comparable agreement ($\kappa = 0.81$), but this value represents a high-confidence subset rather than the full dataset. Our $A_1$--$A_2$ pair achieves $\kappa = 0.8743$ across \textit{all} instances without subsetting.

This achievement is considered notable given that Yor\`{u}b\'{a} is a tonal language, a low-resource language, and our dataset includes naturally occurring examples from diverse sources rather than a single domain.

\section{Conclusion and Future Directions}\label{sec5:conclusion}
This study presents the initial efforts to accelerate the growth of sarcasm detection in low-resource African languages by introducing \textit{Yor-Sac}, a gold-standard manually annotated sarcasm detection dataset in Yor\`{u}b\'{a} language. This language is spoken by over $50$ million people globally, which includes major parts of West Africa and notable southern American countries. 

Sarcasm annotation is particularly challenging, particularly because it requires pragmatic inference beyond literal meaning, alongside cultural and contextual background \citep{lee2025pragmatic}. Within this context, $\kappa = 0.7660$ represents \textit{excellent quality}; moreover, benchmarking (Table~\ref{tab:benchmark_comparison}) demonstrates that we exceed all published English sarcasm annotation studies, including those in high-resource settings with experienced annotators. The achievement of almost perfect agreement ($\kappa = 0.8743$) for one pair represents the highest reported for sarcasm annotation across multiple languages. This indicates the veracity of our resource for steering the progress of sarcasm detection among low-resource African languages.

As \citep{artstein2008inter} emphasizes, agreement expectations are fundamentally task-dependent: objective tasks like part-of-speech tagging routinely achieve $\kappa > 0.95$, while subjective semantic tasks rarely exceed $\kappa = 0.75$. Thus, our substantial inter-annotator agreement (Fleiss' $\kappa = 0.7660$) demonstrates that even for semantically complex phenomena requiring pragmatic inference, native speakers with a shared cultural background in the language can reliably establish subtle distinctions in intended meaning.

Moreover, our agreement analysis provides insights into the semantic structure of sarcasm, which functions as a non-literal linguistic expression. The high unanimity rate $(83.3\%)$ indicates that sarcasm in Yor\`{u}b\'{a}, like in other languages, exhibits markers, such as lexical hyperbole, that native speakers reliably recognize. 

The $16.7\%$ of majority-agreement cases represent instances where sarcasm markers are present, but annotators differ in how much evidence they require before concluding an utterance is sarcastic. This aligns with theoretical accounts of sarcasm as involving scalar rather than categorical semantic interpretation \citep{giora2005irony, wilson2006relevance}. Our soft labels preserve this scalar nature rather than imposing artificial binary categorization. 

Ultimately, traditional $\kappa$ metrics provide only a partial perspective on annotation quality. Thus, by ensuring agreement analysis, our $83.3\%$ unanimity rate reveals that most instances are unambiguous, with the $16.7\%$ variation concentrated on genuinely difficult borderline cases. Recent work on disagreement-aware NLP \citep{uma2021learning, plank-2022-problem} argues that such variation should be preserved as a training signal rather than eliminated, supporting our soft label approach. Future work will build on \textit{Yor-Sarc} to expand the corpus and carry out advanced sarcasm detection experiments.

\section*{Limitation}

Considering the context of learning figurative expressions in the presence of small data, the only observed limitation is the domain coverage of the dataset. This being that our instances span social media and news media, but do not exhaustively cover all Yor\`{u}b\'{a} discourse contexts, such as face-to-face conversation or dialogue. Thus, expanding domain coverage while maintaining agreement quality could be a worthwhile consideration.
Despite the limitation, our comprehensive agreement analysis demonstrates that Yor-Sarc represents a high-quality, reliable resource for computational sarcasm research in African languages.


\noindent
\section*{Acknowledgements}
This publication has emanated from research supported in part by a grant from Taighde \'{E}ireann – Research Ireland under Grant number 18/CRT/6049. For Open Access, the author has applied a CC BY public copyright licence to any Author Accepted Manuscript version arising from this submission.


\section*{Ethics statement}

This research was conducted in accordance with ethical guidelines for human subjects research. All publicly available data (BBC News Yor\`{u}b\'{a}, social media posts) were collected from sources where users had consented to public distribution, and we ensured compliance with platform terms of service. For crowdsourced examples, participants provided informed consent through an ethically approved online survey protocol, with explicit permission to use their contributed examples for research purposes. All three annotators were compensated fairly above minimum wage standards, worked under voluntary agreements with the right to withdraw at any time, and also received comprehensive training on annotation guidelines. To protect privacy, all instances in the dataset have been anonymized. Moreover, the dataset is to be released under a Creative Commons license for research purposes and development of NLP tools for African languages. Ultimately, we acknowledge that sarcasm detection technology could potentially be misused for censorship or manipulation; thus, we advocate for responsible use focused on improving communication technologies and cultural understanding.

\newpage
\renewcommand{\bibname}{the_references}
\bibliographystyle{plainnat}
\bibliography{the_references}

\begin{thebibliography}{32}
\providecommand{\natexlab}[1]{#1}
\providecommand{\url}[1]{\texttt{#1}}
\expandafter\ifx\csname urlstyle\endcsname\relax
  \providecommand{\doi}[1]{doi: #1}\else
  \providecommand{\doi}{doi: \begingroup \urlstyle{rm}\Url}\fi

\bibitem[Adelani et~al.(2022)Adelani, Neubig, Ruder, Rijhwani, Beukman, Palen-Michel, Lignos, Alabi, Muhammad, Nabende, Dione, Bukula, Mabuya, Dossou, Sibanda, Buzaaba, Mukiibi, Kalipe, Mbaye, Taylor, Kabore, Emezue, Aremu, Ogayo, Gitau, Munkoh-Buabeng, Memdjokam~Koagne, Tapo, Macucwa, Marivate, Mboning, Gwadabe, Adewumi, Ahia, Nakatumba-Nabende, Mokono, Ezeani, Chukwuneke, Adeyemi, Hacheme, Abdulmumin, Ogundepo, Yousuf, Moteu~Ngoli, and Klakow]{adelani-etal-2022-masakhaner}
David~Ifeoluwa Adelani, Graham Neubig, Sebastian Ruder, Shruti Rijhwani, Michael Beukman, Chester Palen-Michel, Constantine Lignos, Jesujoba~O. Alabi, Shamsuddeen~H. Muhammad, Peter Nabende, Cheikh M.~Bamba Dione, Andiswa Bukula, Rooweither Mabuya, Bonaventure F.~P. Dossou, Blessing Sibanda, Happy Buzaaba, Jonathan Mukiibi, Godson Kalipe, Derguene Mbaye, Amelia Taylor, Fatoumata Kabore, Chris~Chinenye Emezue, Anuoluwapo Aremu, Perez Ogayo, Catherine Gitau, Edwin Munkoh-Buabeng, Victoire Memdjokam~Koagne, Allahsera~Auguste Tapo, Tebogo Macucwa, Vukosi Marivate, Elvis Mboning, Tajuddeen Gwadabe, Tosin Adewumi, Orevaoghene Ahia, Joyce Nakatumba-Nabende, Neo~L. Mokono, Ignatius Ezeani, Chiamaka Chukwuneke, Mofetoluwa Adeyemi, Gilles~Q. Hacheme, Idris Abdulmumin, Odunayo Ogundepo, Oreen Yousuf, Tatiana Moteu~Ngoli, and Dietrich Klakow.
\newblock {M}asakha{NER} 2.0: {A}frica-centric transfer learning for named entity recognition.
\newblock In Yoav Goldberg, Zornitsa Kozareva, and Yue Zhang, editors, \emph{Proceedings of the 2022 Conference on Empirical Methods in Natural Language Processing}, pages 4488--4508, Abu Dhabi, United Arab Emirates, December 2022. Association for Computational Linguistics.
\newblock \doi{10.18653/v1/2022.emnlp-main.298}.
\newblock URL \url{https://aclanthology.org/2022.emnlp-main.298/}.

\bibitem[Artstein and Poesio(2008)]{artstein2008inter}
Ron Artstein and Massimo Poesio.
\newblock Inter-coder agreement for computational linguistics.
\newblock \emph{Computational linguistics}, 34\penalty0 (4):\penalty0 555--596, 2008.

\bibitem[Asubiaro et~al.(2018)Asubiaro, Adegbola, Mercer, and Ajiferuke]{asubiaro2018word_language_identification}
Toluwase Asubiaro, Tunde Adegbola, Robert Mercer, and Isola Ajiferuke.
\newblock A word-level language identification strategy for resource-scarce languages.
\newblock \emph{Proceedings of the Association for Information Science and Technology}, 55\penalty0 (1):\penalty0 19--28, 2018.

\bibitem[Bamman and Smith(2015)]{bamman2015contextualized}
David Bamman and Noah~A. Smith.
\newblock Contextualized sarcasm detection on twitter.
\newblock In \emph{Proceedings of the International AAAI Conference on Web and Social Media}, volume~9, pages 574--577, 2015.
\newblock URL \url{https://ojs.aaai.org/index.php/ICWSM/article/view/14655}.

\bibitem[Castro et~al.(2019)Castro, Hazarika, P{\'e}rez-Rosas, Zimmermann, Mihalcea, and Poria]{castro-etal-2019-towards}
Santiago Castro, Devamanyu Hazarika, Ver{\'o}nica P{\'e}rez-Rosas, Roger Zimmermann, Rada Mihalcea, and Soujanya Poria.
\newblock Towards multimodal sarcasm detection (an {\_}{O}bviously{\_} perfect paper).
\newblock In Anna Korhonen, David Traum, and Llu{\'i}s M{\`a}rquez, editors, \emph{Proceedings of the 57th Annual Meeting of the Association for Computational Linguistics}, pages 4619--4629, Florence, Italy, July 2019. Association for Computational Linguistics.
\newblock \doi{10.18653/v1/P19-1455}.
\newblock URL \url{https://aclanthology.org/P19-1455/}.

\bibitem[Davidov et~al.(2010)Davidov, Tsur, and Rappoport]{davidov2010semi}
Dmitry Davidov, Oren Tsur, and Ari Rappoport.
\newblock Semi-supervised recognition of sarcasm in twitter and amazon.
\newblock In \emph{Proceedings of the fourteenth conference on computational natural language learning}, pages 107--116, 2010.

\bibitem[Dione et~al.(2023)Dione, Adelani, Nabende, Alabi, Sindane, Buzaaba, Muhammad, Emezue, Ogayo, Aremu, et~al.]{dione2023masakhapos}
Cheikh M~Bamba Dione, David~Ifeoluwa Adelani, Peter Nabende, Jesujoba Alabi, Thapelo Sindane, Happy Buzaaba, Shamsuddeen~Hassan Muhammad, Chris~Chinenye Emezue, Perez Ogayo, Anuoluwapo Aremu, et~al.
\newblock Masakhapos: Part-of-speech tagging for typologically diverse african languages.
\newblock In \emph{Proceedings of the 61st Annual Meeting of the Association for Computational Linguistics (Volume 1: Long Papers)}, pages 10883--10900, 2023.

\bibitem[Efron and Tibshirani(1985)]{efron1985bootstrap}
Bradley Efron and Robert Tibshirani.
\newblock The bootstrap method for assessing statistical accuracy.
\newblock \emph{Behaviormetrika}, 12\penalty0 (17):\penalty0 1--35, 1985.

\bibitem[Fagbolu et~al.(2016)Fagbolu, Obalalu, Udoh, and Uyo]{fagbolu2016applying}
Olutola~Olaide Fagbolu, Babatunde~Sunday Obalalu, Samuel~S Udoh, and Ibadan~Abeokuta Uyo.
\newblock Applying rough set theory to yorub{\'a} language translation.
\newblock In \emph{International Conference on Advanced Trends in ICT and Management (ICAITM) 28th}, 2016.

\bibitem[Farha and Magdy(2020)]{farha2020arabic}
Ibrahim~Abu Farha and Walid Magdy.
\newblock From arabic sentiment analysis to sarcasm detection: The arsarcasm dataset.
\newblock In \emph{The 4th Workshop on Open-Source Arabic Corpora and Processing Tools}, pages 32--39. European Language Resources Association (ELRA), 2020.

\bibitem[Fleiss(1971)]{fleiss1971measuring}
Joseph~L. Fleiss.
\newblock Measuring nominal scale agreement among many raters.
\newblock \emph{Psychological Bulletin}, 76\penalty0 (5):\penalty0 378--382, 1971.
\newblock \doi{10.1037/h0031619}.

\bibitem[Gao et~al.(2024)Gao, Nayak, and Coler]{gao2024sarcasm}
Xiyuan Gao, Shekhar Nayak, and Matt Coler.
\newblock Improving sarcasm detection from speech and text through attention-based fusion exploiting the interplay of emotions and sentiments.
\newblock In \emph{Proceedings of Meetings on Acoustics}, volume~54. Acoustical Society of America, 2024.

\bibitem[Giora et~al.(2005)Giora, Federman, Kehat, Fein, and Sabah]{giora2005irony}
Rachel Giora, Shani Federman, Arnon Kehat, Ofer Fein, and Hadas Sabah.
\newblock Irony aptness.
\newblock \emph{Humor: International Journal of Humor Research}, 18\penalty0 (1), 2005.

\bibitem[Gonz{\'a}lez-Ib{\'a}nez et~al.(2011)Gonz{\'a}lez-Ib{\'a}nez, Muresan, and Wacholder]{gonzalez2011identifying}
Roberto Gonz{\'a}lez-Ib{\'a}nez, Smaranda Muresan, and Nina Wacholder.
\newblock Identifying sarcasm in twitter: a closer look.
\newblock In \emph{Proceedings of the 49th annual meeting of the association for computational linguistics: human language technologies}, pages 581--586, 2011.

\bibitem[Jimoh et~al.(2025)Jimoh, {De Wille}, and Nikolov]{jimohslr}
Toheeb~Aduramomi Jimoh, Tabea {De Wille}, and Nikola~S. Nikolov.
\newblock Bridging gaps in natural language processing for yor\`{u}b\'{a}': A systematic review of a decade of progress and prospects.
\newblock \emph{Natural Language Processing Journal}, 13:\penalty0 100194, 2025.
\newblock ISSN 2949-7191.
\newblock \doi{https://doi.org/10.1016/j.nlp.2025.100194}.
\newblock URL \url{https://www.sciencedirect.com/science/article/pii/S2949719125000706}.

\bibitem[Joshi et~al.(2017)Joshi, Bhattacharyya, and Carman]{joshi2017automatic}
Aditya Joshi, Pushpak Bhattacharyya, and Mark~J Carman.
\newblock Automatic sarcasm detection: A survey.
\newblock \emph{ACM Computing Surveys (CSUR)}, 50\penalty0 (5):\penalty0 1--22, 2017.

\bibitem[Kamau and Abbas(2025)]{kamau2025multimodal_swahili}
Eugene~Kariba Kamau and Noorhan Abbas.
\newblock Multimodal sarcasm dataset generation for a low-resource language: Swahili.
\newblock In \emph{International Conference on Innovative Techniques and Applications of Artificial Intelligence}, pages 239--252. Springer, 2025.

\bibitem[Khan et~al.(2024)Khan, Qasim, Khan, Khan, Ali~Khan, Qahmash, and Ghadi]{khan2024automated_urdu}
Shumaila Khan, Iqbal Qasim, Wahab Khan, Aurangzeb Khan, Javed Ali~Khan, Ayman Qahmash, and Yazeed~Yasin Ghadi.
\newblock An automated approach to identify sarcasm in low-resource language.
\newblock \emph{PloS one}, 19\penalty0 (12):\penalty0 e0307186, 2024.

\bibitem[Khodak et~al.(2018)Khodak, Saunshi, and Vodrahalli]{khodak2018large_SARC_dataset}
Mikhail Khodak, Nikunj Saunshi, and Kiran Vodrahalli.
\newblock A large self-annotated corpus for sarcasm.
\newblock In \emph{proceedings of the eleventh international conference on language resources and evaluation (LREC 2018)}, 2018.

\bibitem[Ladoja and Afape(2024)]{ladoja2024sarcasm}
Khadijat~T Ladoja and Ruth~T Afape.
\newblock Sarcasm detection in pidgin tweets using machine learning techniques.
\newblock \emph{Asian Journal of Research in Computer Science}, 17\penalty0 (5):\penalty0 212--221, 2024.

\bibitem[Landis and Koch(1977)]{landis1977measurement}
J.~Richard Landis and Gary~G. Koch.
\newblock The measurement of observer agreement for categorical data.
\newblock \emph{Biometrics}, 33\penalty0 (1):\penalty0 159--174, 1977.
\newblock \doi{10.2307/2529310}.

\bibitem[Lee et~al.(2025)Lee, Fong, Le, Shah, Han, and Zhu]{lee2025pragmatic}
Joshua Lee, Wyatt Fong, Alexander Le, Sur Shah, Kevin Han, and Kevin Zhu.
\newblock Pragmatic metacognitive prompting improves llm performance on sarcasm detection.
\newblock In \emph{Proceedings of the 1st Workshop on Computational Humor (CHum)}, pages 63--70, 2025.

\bibitem[McHugh(2012)]{mchugh2012interrater_kappa}
Mary~L McHugh.
\newblock Interrater reliability: the kappa statistic.
\newblock \emph{Biochemia medica}, 22\penalty0 (3):\penalty0 276--282, 2012.

\bibitem[Misra and Arora(2023)]{misra2023sarcasm}
Rishabh Misra and Prahal Arora.
\newblock Sarcasm detection using news headlines dataset.
\newblock \emph{AI Open}, 4:\penalty0 13--18, 2023.

\bibitem[Muhammad et~al.(2022)Muhammad, Adelani, Ruder, Ahmad, Abdulmumin, Bello, Choudhury, Emezue, Abdullahi, Aremu, et~al.]{muhammad2022naijasenti}
Shamsuddeen~Hassan Muhammad, David~Ifeoluwa Adelani, Sebastian Ruder, Ibrahim~Sa’id Ahmad, Idris Abdulmumin, Bello~Shehu Bello, Monojit Choudhury, Chris~Chinenye Emezue, Saheed~Salahudeen Abdullahi, Anuoluwapo Aremu, et~al.
\newblock Naijasenti: A nigerian twitter sentiment corpus for multilingual sentiment analysis.
\newblock In \emph{Proceedings of the Thirteenth Language Resources and Evaluation Conference}, pages 590--602, 2022.

\bibitem[Muhammad et~al.(2023)Muhammad, Abdulmumin, Ayele, Ousidhoum, Adelani, Yimam, Ahmad, Beloucif, Mohammad, Ruder, et~al.]{muhammad2023afrisenti}
Shamsuddeen~Hassan Muhammad, Idris Abdulmumin, Abinew~Ali Ayele, Nedjma Ousidhoum, David~Ifeoluwa Adelani, Seid~Muhie Yimam, Ibrahim~Sa'id Ahmad, Meriem Beloucif, Saif Mohammad, Sebastian Ruder, et~al.
\newblock Afrisenti: A twitter sentiment analysis benchmark for african languages.
\newblock In \emph{Proceedings of the 2023 conference on empirical methods in natural language processing}, pages 13968--13981, 2023.

\bibitem[Plank(2022)]{plank-2022-problem}
Barbara Plank.
\newblock The ``problem'' of human label variation: On ground truth in data, modeling and evaluation.
\newblock In Yoav Goldberg, Zornitsa Kozareva, and Yue Zhang, editors, \emph{Proceedings of the 2022 Conference on Empirical Methods in Natural Language Processing}, pages 10671--10682, Abu Dhabi, United Arab Emirates, December 2022. Association for Computational Linguistics.
\newblock \doi{10.18653/v1/2022.emnlp-main.731}.
\newblock URL \url{https://aclanthology.org/2022.emnlp-main.731/}.

\bibitem[Riloff et~al.(2013)Riloff, Qadir, Surve, De~Silva, Gilbert, and Huang]{riloff-etal-2013-sarcasm}
Ellen Riloff, Ashequl Qadir, Prafulla Surve, Lalindra De~Silva, Nathan Gilbert, and Ruihong Huang.
\newblock Sarcasm as contrast between a positive sentiment and negative situation.
\newblock In David Yarowsky, Timothy Baldwin, Anna Korhonen, Karen Livescu, and Steven Bethard, editors, \emph{Proceedings of the 2013 Conference on Empirical Methods in Natural Language Processing}, pages 704--714, Seattle, Washington, USA, October 2013. Association for Computational Linguistics.
\newblock URL \url{https://aclanthology.org/D13-1066/}.

\bibitem[Swami et~al.(2018)Swami, Khandelwal, Singh, Akhtar, and Shrivastava]{swami2018corpus}
Sahil Swami, Ankush Khandelwal, Vinay Singh, Syed~Sarfaraz Akhtar, and Manish Shrivastava.
\newblock A corpus of english-hindi code-mixed tweets for sarcasm detection.
\newblock \emph{arXiv preprint arXiv:1805.11869}, 2018.

\bibitem[Uma et~al.(2021)Uma, Fornaciari, Hovy, Paun, Plank, and Poesio]{uma2021learning}
Alexandra~N Uma, Tommaso Fornaciari, Dirk Hovy, Silviu Paun, Barbara Plank, and Massimo Poesio.
\newblock Learning from disagreement: A survey.
\newblock \emph{Journal of Artificial Intelligence Research}, 72:\penalty0 1385--1470, 2021.

\bibitem[Wilson and Wharton(2006)]{wilson2006relevance}
Deirdre Wilson and Tim Wharton.
\newblock Relevance and prosody.
\newblock \emph{Journal of pragmatics}, 38\penalty0 (10):\penalty0 1559--1579, 2006.

\bibitem[Zhang et~al.(2025)Zhang, Zou, Lian, Tiwari, and Qin]{zhang2025sarcasmbench}
Yazhou Zhang, Chunwang Zou, Zheng Lian, Prayag Tiwari, and Jing Qin.
\newblock Sarcasmbench: Towards evaluating large language models on sarcasm understanding.
\newblock \emph{IEEE Transactions on Affective Computing}, 2025.

\end{thebibliography}

\appendix

\newpage
\section{Supplementary Agreement Analysis}
\label{sec:appendix}

\subsection{Label Distribution Across Annotators}

\begin{figure}[htbp]
\centering
\includegraphics[width=0.8\textwidth]{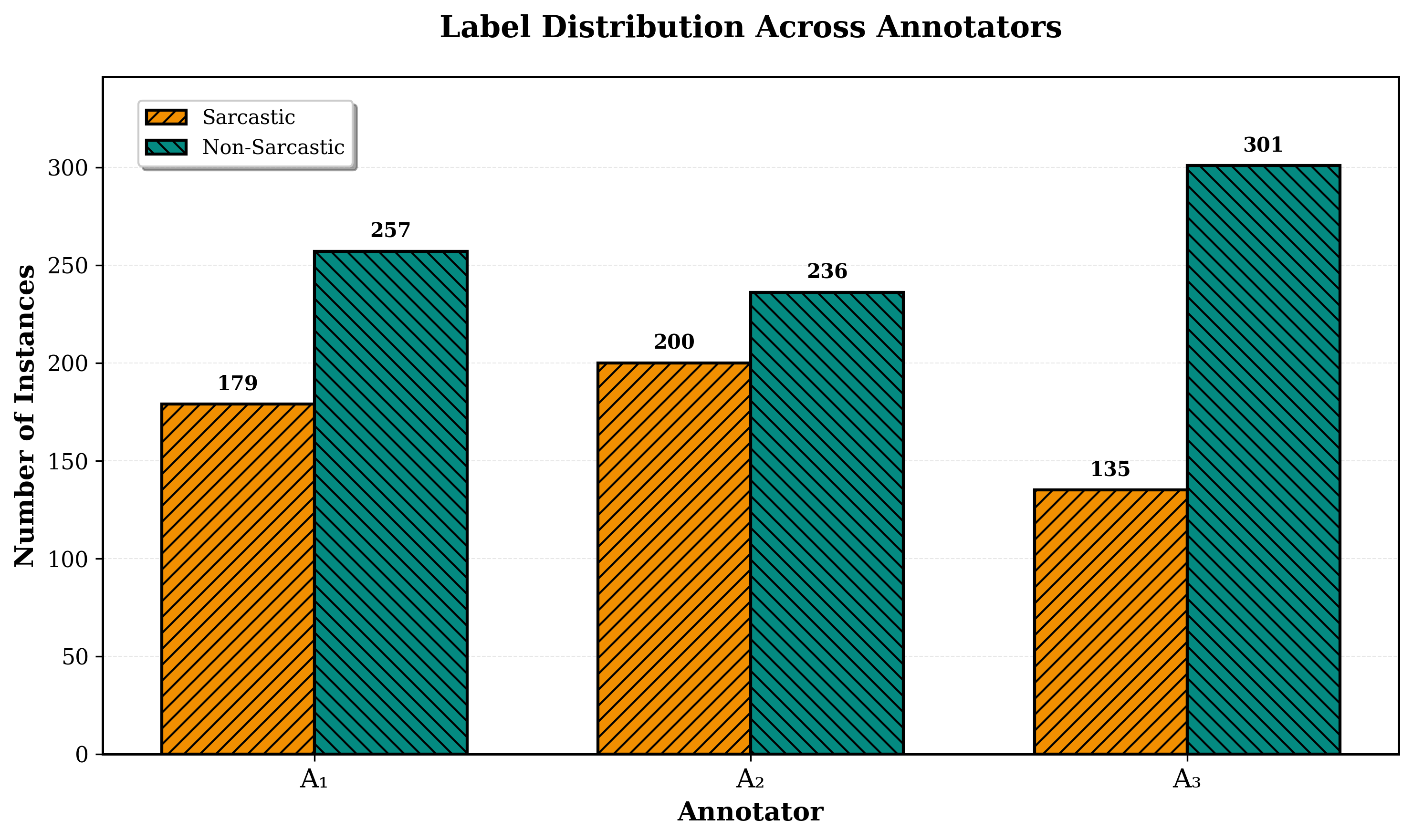}
\caption{Label distribution across annotators}
\label{fig:appendix_annotator_dist}
\end{figure}

Figure~\ref{fig:appendix_annotator_dist} shows label distributions for each annotator. Sarcastic label rates range from $30.96\%$ ($A_3$) to $45.87\%$ ($A_2$), reflecting threshold differences while maintaining balanced class distributions. 

\subsection{Annotator Labelling Bias}

\begin{figure}[htbp]
\centering
\includegraphics[width=0.8\textwidth]{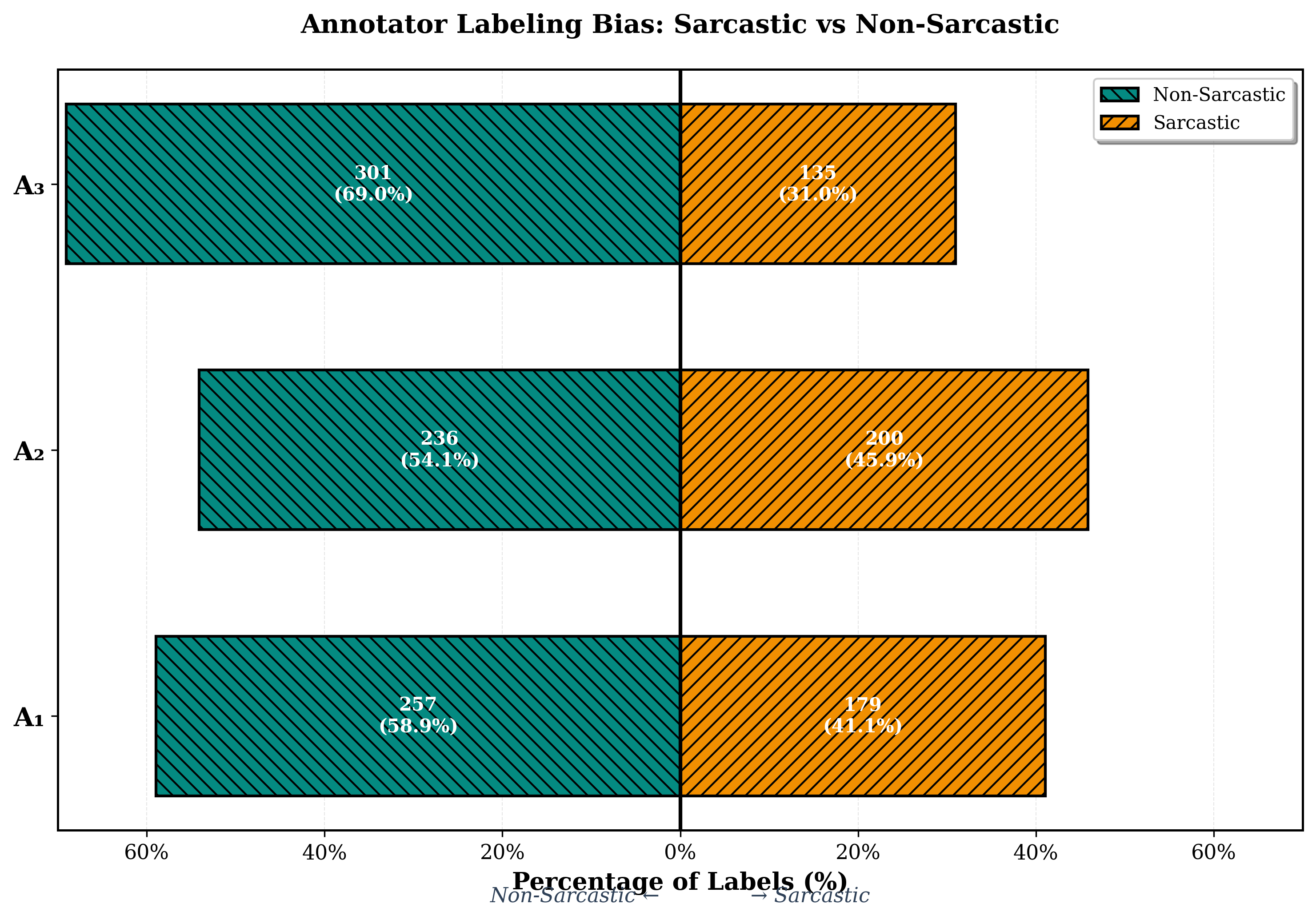}
\caption{Annotator Labelling bias}
\label{fig:appendix_bias}
\end{figure}

Figure~\ref{fig:appendix_bias} illustrates threshold variation: $A_2$ most liberal $(45.87\%)$, $A_3$ most conservative $(30.96\%)$. All annotators maintain balanced distributions without extreme bias. The threshold differences explain pairwise agreement patterns while all maintain balance.

\subsection{Agreement Level Distribution}

\begin{figure}[htbp]
\centering
\includegraphics[width=0.95\textwidth]{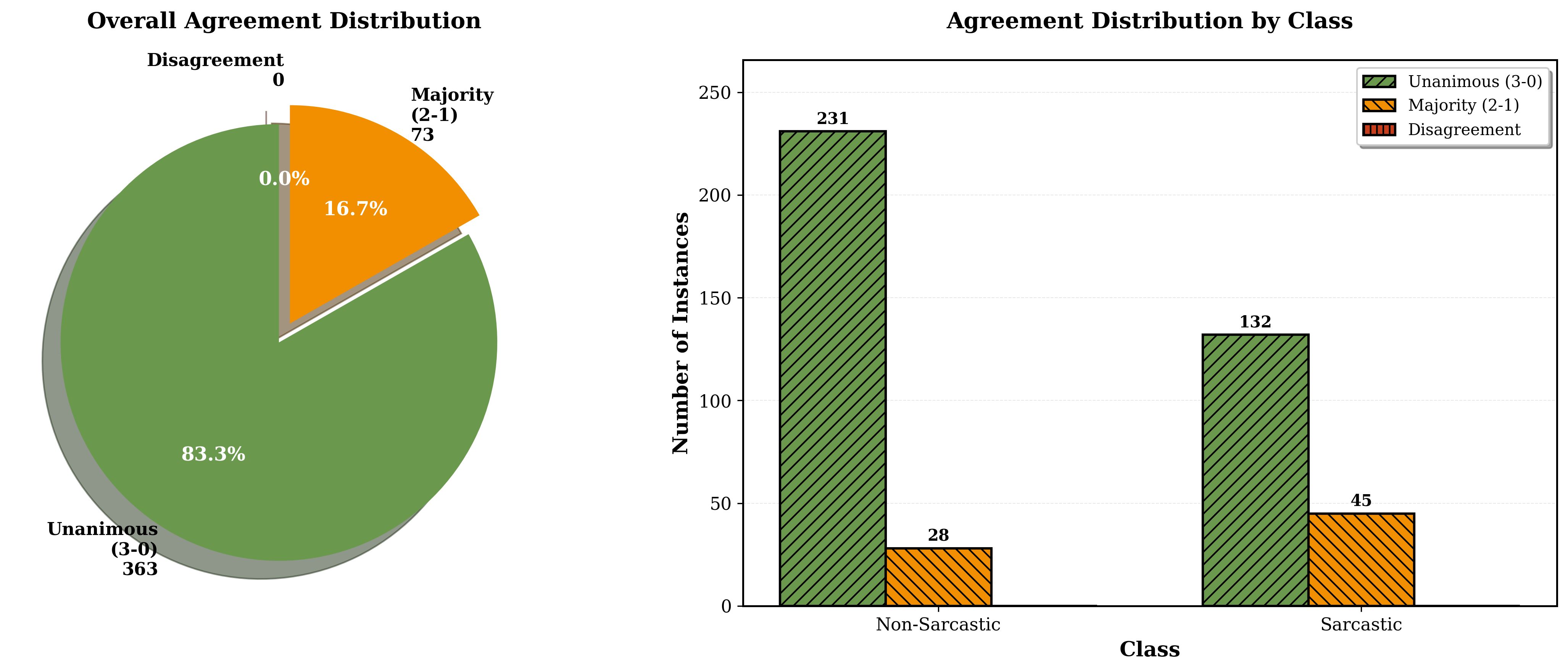}
\caption{Overall agreement distribution (left) and distribution by class (right)}
\label{fig:appendix_agreement_dist}
\end{figure}

Figure~\ref{fig:appendix_agreement_dist} reveals $83.3\%$ unanimity with zero complete disagreements---as required. Sarcastic instances show slightly higher consensus $(85.5\%)$ than non-sarcastic $(81.5\%)$.

\addcontentsline{toc}{chapter}{References}
\end{document}